%% file: main.tex
\documentclass{article} 
\usepackage{iclr2026_conference,times}

\usepackage{natbib}   
\input{math_commands.tex}

\usepackage{adjustbox}
\usepackage{hyperref}
\usepackage{url}
\usepackage{graphicx} 
\usepackage{subcaption}
\usepackage{tikz}
\usetikzlibrary{shapes.geometric, arrows.meta, positioning, calc, shadows.blur}
\usepackage{xcolor}
\usepackage{amssymb} 
\usepackage[utf8]{inputenc}
\usepackage[T1]{fontenc}
\usepackage{booktabs}
\usepackage{amsmath, graphicx}
\usepackage{amsthm}
\usepackage{enumitem} 
\usepackage{multicol}
\usepackage{tkz-euclide}
\usepackage{tikz}
\usetikzlibrary{arrows.meta, positioning}
\usepackage{wrapfig}
\usepackage{longtable}
\usepackage{forest}
\usepackage{amsfonts}
\usepackage{algorithm}
\usepackage{xcolor}
\definecolor{lightgray}{gray}{0.9}
\usepackage{algpseudocode}
\usetikzlibrary{positioning}
\usepackage{skull}
\usetikzlibrary{positioning,arrows.meta,shapes.multipart,calc}
\usepackage{colortbl}
\usepackage{float}
\usepackage{amstext}
\usepackage{array}
\newcolumntype{L}{>{$}l<{$}} 

\usepackage[table]{xcolor}
\usepackage{booktabs}
\usepackage{xcolor}
\usepackage{colortbl}
\usepackage{float}
\definecolor{Red}{rgb}{1,0,0}
\definecolor{Blue}{rgb}{0,0,1}
\definecolor{Purple}{rgb}{.5,0,.5}
\definecolor{Orange}{rgb}{1,.35,0}

\definecolor{RoyalBlue}{cmyk}{1, 0.50, 0, 0}

\definecolor{myyellow}{RGB}{204, 153, 0}
\usepackage{placeins}
\usepackage{hyperref}
\usepackage[capitalize]{cleveref}

\title{When Stability Fails: Hidden Failure Modes of LLMs in Data-Constrained Scientific Decision-Making}

\author{Nazia Riasat\\
North Dakota State University\\
\texttt{nazia.riasat@ndsu.edu}
}

\begin{document}
\iclrfinalcopy
\maketitle
\begin{abstract}

Large language models (LLMs) are increasingly used as decision-support tools in data-constrained scientific workflows, where correctness and validity are critical. However, evaluation practices often emphasize stability or reproducibility across repeated runs. While these properties are desirable, stability alone does not guarantee agreement with statistical ground truth when such references are available. We introduce a controlled behavioral evaluation framework that explicitly separates four dimensions of LLM decision-making: stability, correctness, prompt sensitivity, and output validity under fixed statistical inputs. We evaluate multiple LLMs using a statistical gene prioritization task derived from differential expression analysis across prompt regimes involving strict and relaxed significance thresholds, borderline ranking scenarios, and minor wording variations. Our experiments show that LLMs can exhibit near-perfect run-to-run stability while systematically diverging from statistical ground truth, over-selecting under relaxed thresholds, responding sharply to minor prompt wording changes, or producing syntactically plausible gene identifiers absent from the input table. Although stability reflects robustness across repeated runs, it does not guarantee agreement with statistical ground truth in structured scientific decision tasks. These findings highlight the importance of explicit ground-truth validation and output validity checks when deploying LLMs in automated or semi-automated scientific workflows.
\end{abstract}

\section{Introduction}
\begin{figure}[t]
\centering
\begin{tikzpicture}[
scale=0.85,
transform shape,
font=\footnotesize\sffamily,
box/.style={
draw,
rounded corners=2pt,
inner sep=6pt,
align=left
},
arrow/.style={-{Latex[length=2mm]}, line width=0.6pt}
]

\definecolor{setupblue}{RGB}{59,130,189}
\definecolor{stabilityorange}{RGB}{237,125,49}
\definecolor{promptpurple}{RGB}{165,81,148}
\definecolor{hallucinred}{RGB}{214,96,77}
\definecolor{summarygreen}{RGB}{91,155,103}

\node[box, text width=0.82\linewidth, fill=setupblue!6, draw=setupblue!60] (A) {%
  \textcolor{setupblue}{\textbf{(A) Experimental setup}}\\
  Same DE table $\rightarrow$ 3 LLMs (ChatGPT, Gemini, Claude), 10 runs\\
  \textit{Regimes:} FDR$\le$0.05, 0.05 $<$FDR$\le$0.10, borderline Top-20, wording (7a vs 7b), ranked Top-20
};

\node[box, text width=0.38\linewidth, anchor=north west, fill=stabilityorange!6, draw=stabilityorange!60]
  (B) at ($ (A.south west) + (0,-5mm) $) {%
  \textcolor{stabilityorange}{\textbf{(B) Stability $\neq$ correctness}}\\
  Identical runs (Jaccard$\approx$1) can still disagree with ground truth.
};

\node[box, text width=0.38\linewidth, anchor=north east, fill=hallucinred!6, draw=hallucinred!60]
  (D) at ($ (A.south east) + (0,-5mm) $) {%
  \textcolor{hallucinred}{\textbf{(D) Hallucinated identifiers}}\\
  Outputs may include gene IDs absent from the input table.
};

\node[box, text width=0.38\linewidth, anchor=north west, fill=promptpurple!6, draw=promptpurple!60]
  (C) at ($ (B.south west) + (0,-4mm) $) {%
  \textcolor{promptpurple}{\textbf{(C) Prompt sensitivity}}\\
  Minor wording changes yield different Top-20 sets.
};

\node[box, text width=0.38\linewidth, anchor=north east, fill=summarygreen!6, draw=summarygreen!60]
  (E) at ($ (D.south east) + (0,-4mm) $) {%
  \textcolor{summarygreen}{\textbf{(E) Failure modes}}\\
  Over-selection, prompt sensitivity, hallucination.
};

\draw[arrow] (A.south west) ++(2cm,0) -- (B.north);
\draw[arrow] (A.south east) ++(-2cm,0) -- (D.north);
\draw[arrow] (B.south) -- (C.north);
\draw[arrow] (D.south) -- (E.north);

\end{tikzpicture}

\vspace{-2mm}
\caption{
Failure modes in LLM-based statistical gene prioritization under fixed input data. Despite high internal stability, models may disagree with the DESeq2-derived statistical reference (B), exhibit prompt sensitivity (C), or produce invalid gene identifiers (D), Panel (E) summarizes observed behaviors.
}
\label{fig:schematic}
\end{figure}
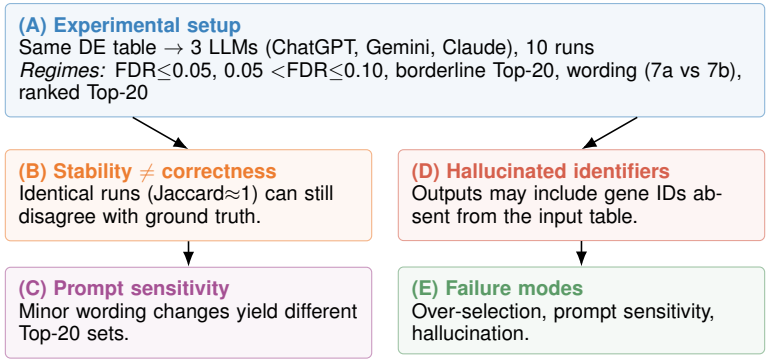

Large language models (LLMs) are increasingly explored as decision-support tools in scientific and biomedical workflows, including tasks such as data interpretation, hypothesis generation, and candidate gene prioritization \citep{Singhal2023, OpenAI2023}. In bioinformatics settings, LLMs are often applied to summarize or rank outputs from statistical analyses, where their fluent reasoning and apparent consistency across repeated runs can encourage trust among users. However, evaluation practices frequently emphasize stability or reproducibility across repeated runs as indicators of reliability. While stability is generally desirable, it does not necessarily imply correctness in structured scientific decision tasks. In many machine learning settings, stability is viewed as an indicator of robustness to perturbations in data or model inputs. However, stability alone cannot substitute for explicit validation when reliable statistical references exist. When the underlying statistical reference is known, stable model outputs may still diverge systematically from statistically valid conclusions. In this work, we investigate this phenomenon through a controlled evaluation of LLM behavior in a statistical gene prioritization task derived from differential expression analysis. In differential expression studies, statistical methods identify genes whose expression differs significantly between experimental conditions. Tools such as DESeq2 estimate gene-level effect sizes and statistical significance while controlling the false discovery rate (FDR) \citep{love2014, benjamini1995}. Here we treat the set of genes identified by a deterministic DESeq2 analysis as a fixed statistical reference that enables controlled comparison of LLM outputs under identical inputs. Using a fixed differential expression table as input, we query multiple LLMs across prompt regimes that vary statistical thresholds, borderline selection scenarios, and prompt wording. We evaluated each configuration over repeated runs to measure stability, agreement with the statistical reference, prompt sensitivity, and output validity. This design isolates model behavior from data variability and enables systematic identification of failure modes that may remain hidden in more open-ended LLM evaluations. Importantly, our analysis does not challenge the value of stability itself, but demonstrates that stability alone can be insufficient for reliable decision-making when ground-truth statistical references are available. Our results reveal several characteristic behaviors. First, models often produce highly stable outputs across repeated runs while exhibiting low agreement with the statistical reference. Second, small wording changes in prompts can alter gene prioritization outcomes, reflecting shifts in how models interpret statistical objectives. Third, relaxed statistical thresholds encourage over-selection relative to the reference procedure. Finally, models may produce syntactically plausible but invalid gene identifiers that do not appear in the input table. Collectively, these results indicate that stability should be interpreted as a complementary diagnostic rather than a proxy for correctness in data-constrained scientific workflows. Figure~\ref{fig:schematic} summarizes the experimental setup and primary failure modes observed in our evaluation. We therefore introduce a controlled behavioral evaluation framework that separates four dimensions of LLM decision-making: stability, correctness relative to a statistical reference, prompt sensitivity, and output validity.

\vspace{-0.3cm}
\section{Experimental Setup and Evaluation}
We study LLM behavior in a controlled statistical gene prioritization task derived from differential expression (DE) analysis. In this setting, genes are ranked according to significance estimates produced by the DESeq2 statistical model. A single fixed differential expression table is provided as input to all models, and no external biological knowledge or contextual information is permitted. Differential expression statistics are computed using standard normalization, dispersion estimation, and multiple-testing correction procedures \citep{ji2023, benjamini1995}. We evaluate three large language models: ChatGPT (GPT-5.2), Google Gemini 3, and Claude Opus 4.5, across several prompt regimes reflecting common analytical scenarios. These include strict statistical thresholding ($\mathrm{FDR} \leq 0.05$), relaxed thresholding ($0.05 < \mathrm{FDR} \leq 0.10$), borderline Top\text{-}20 selection, and ranked Top\text{-}20 prioritization. To assess prompt sensitivity, we construct two semantically equivalent prompts that differ only in wording (P7a vs.\ P7b). Although both prompts encode the same decision criteria, minor differences in emphasis (e.g., statistical significance versus ranking instructions) may introduce subtle shifts in task interpretation. Each configuration is evaluated over ten repeated runs with identical inputs. Model behavior is assessed along four complementary evaluation dimensions. First, we measure run-to-run stability of model outputs using the Jaccard index \citep{jaccard1901}, which quantifies similarity between gene sets produced across repeated executions. Because predicted gene sets may vary in size across prompt regimes, we additionally report the overlap coefficient (Szymkiewicz–Simpson coefficient) \citep{vijaymeena2016}, a containment-based similarity measure that captures cases where one set is largely included within another. To evaluate correctness, we compare model outputs against a deterministic statistical reference derived from DESeq2 differential expression analysis \citep{love2014}. Prompt sensitivity is quantified by measuring divergence between outputs generated from semantically equivalent prompts \citep{zhu2023}. Finally, we examine output validity by identifying gene identifiers produced by the model that do not appear in the input table, reflecting hallucination-related failure modes previously documented in language models \citep{li2024, kaddour2023}. In this study, the DESeq2 output serves as a deterministic statistical reference that enables controlled comparisons across model outputs. While borderline FDR regimes may introduce ambiguity in biological interpretation, the reference functions here as a consistent evaluation baseline rather than a claim of absolute biological ground truth. Code and experimental outputs are provided in the supplementary repository described in Appendix H4.

\section{Results}
Failure Modes in LLM-Based Gene Prioritization: We organize the results by failure mode rather than by individual prompt so that stability, agreement with the DESeq2-derived statistical reference, prompt sensitivity, and output validity can be compared directly under identical inputs.

\subsection{Stability Does Not Imply Correctness across LLMs}

\begin{figure}[htbp]
  \centering
  \includegraphics[width=0.62\linewidth]{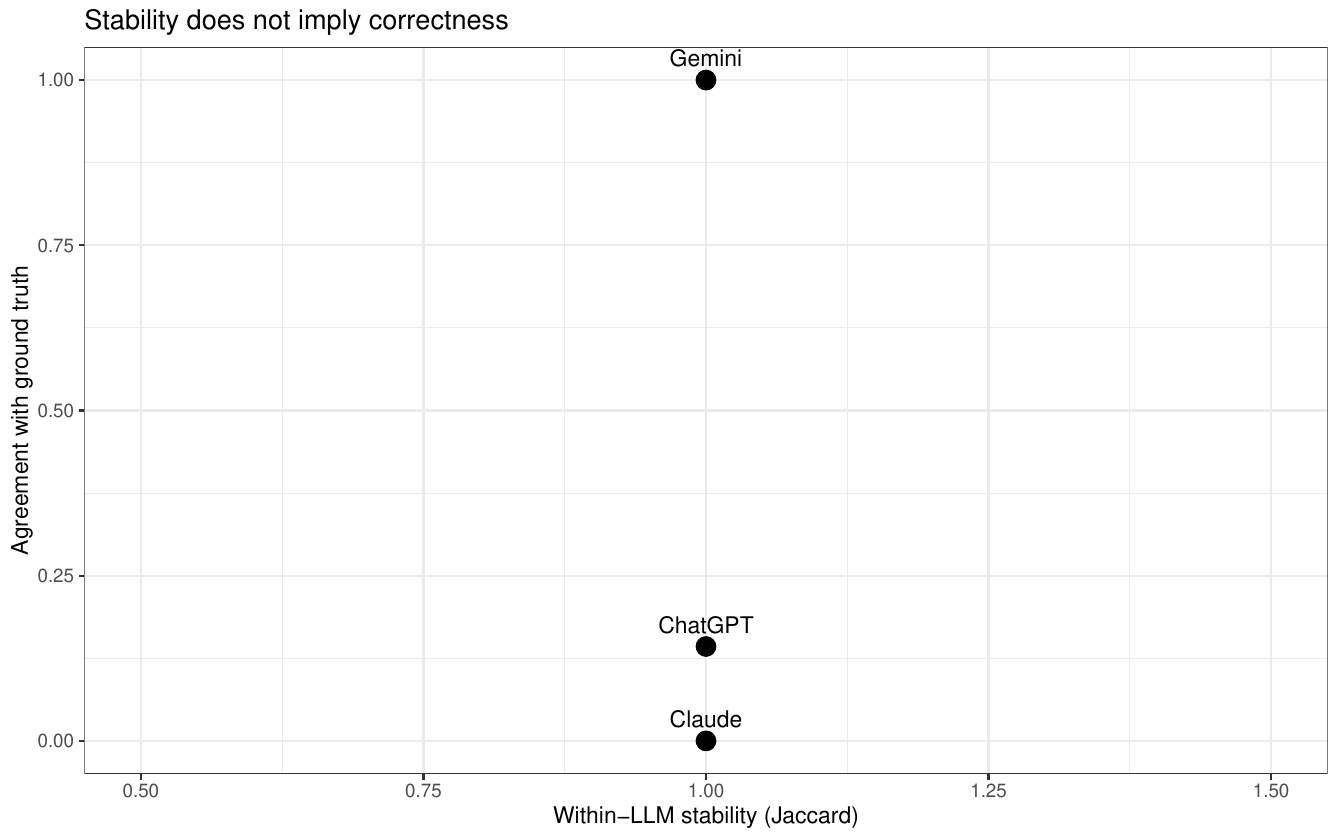}
  \caption{Stability does not imply correctness. Within-LLM stability (pairwise Jaccard similarity across repeated runs; x-axis) versus agreement with the DESeq2-derived statistical reference (Jaccard against ground truth; y-axis). Each point represents the mean stability and agreement values aggregated across 10 repeated runs per configuration. Run-level variability was minimal due to near-perfect within-model stability.}
  \label{fig:stability}
\end{figure}

Across all evaluated prompt regimes, the three large language models exhibit near-perfect run-to-run stability. In threshold-based tasks (e.g., $\mathrm{FDR} \leq 0.05$ and $0.05 < \mathrm{FDR} \leq 0.10$), repeated queries under fixed inputs typically produce identical outputs. However, while this apparent determinism reflects high internal stability, it does not imply correctness. When compared against the DESeq2-derived statistical reference, stable outputs may still show low or even zero agreement with the statistical reference. A Jaccard value of zero indicates that the predicted gene set shares no identifiers with the reference set; this may arise either from returning an empty set or from selecting genes entirely outside the ground-truth set. As shown in Figures~\ref{fig:stability} and~\ref{fig:prompt_regimes_sensitivity}, models achieve maximal within-LLM stability while exhibiting substantial variation in agreement with the reference. Under relaxed thresholds ($0.05 < \mathrm{FDR} \leq 0.10$), corresponding to the borderline discovery regime described in Section 2, some models consistently return large gene sets that only partially overlap with DESeq2-significant genes, while others return incorrect or empty selections. These results demonstrate that run-to-run consistency alone is insufficient as evidence of correct statistical reasoning.

\subsection{Prompt Regime and Wording Sensitivity under identical inputs}

\begin{figure}[!ht]
    \centering
    \includegraphics[width=0.55\linewidth]{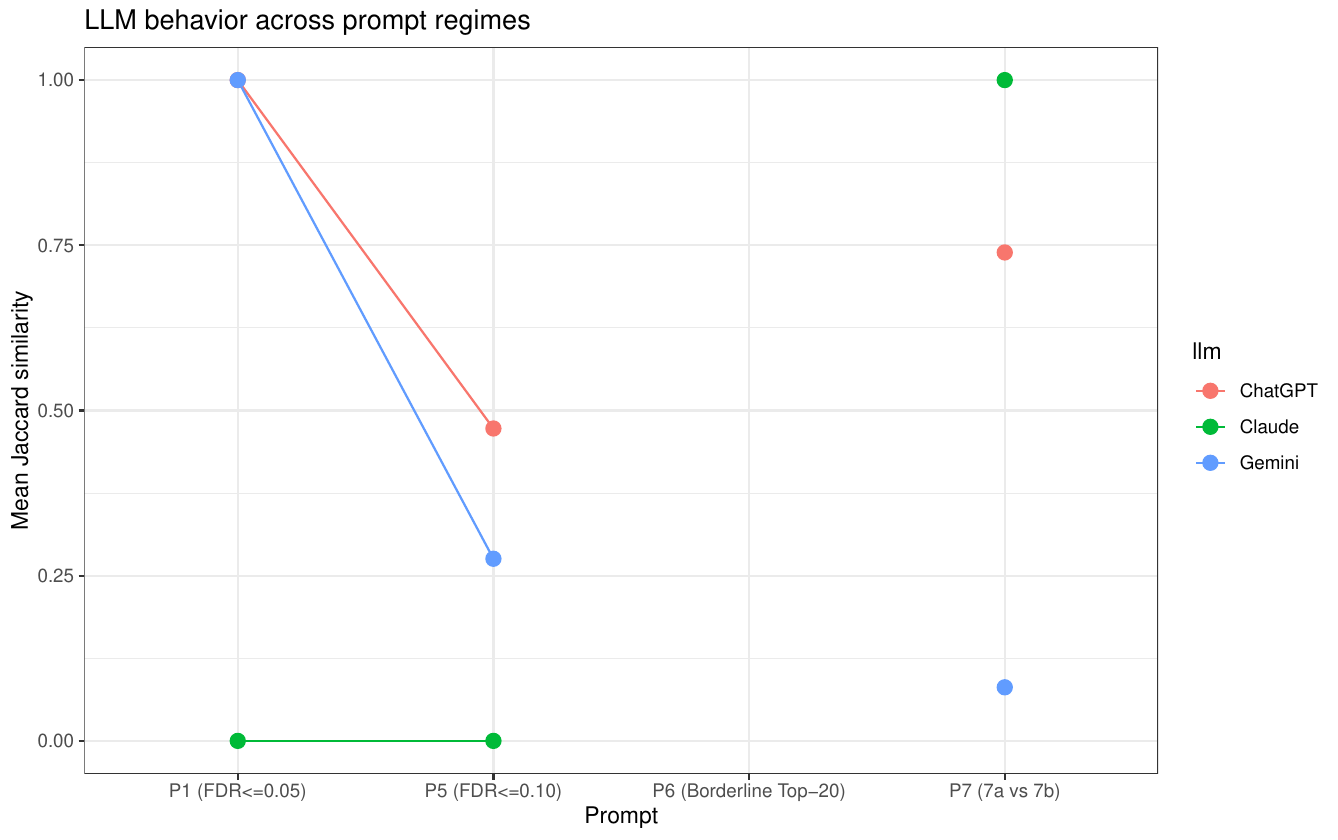}
    \caption{LLM behavior across prompt regimes. Mean Jaccard similarity to DESeq2 reference of Top-20 gene sets across threshold-based selection, borderline prioritization, and prompt wording variants.}
    \label{fig:prompt_regimes_sensitivity}
\end{figure}

In addition to the Jaccard index, we report the overlap coefficient to capture containment relationships between gene sets of different sizes. To assess sensitivity to prompt formulation, we evaluated model outputs across multiple prompt regimes while holding the underlying differential expression table fixed. Figure~\ref{fig:prompt_regimes_sensitivity} summarizes mean Jaccard similarity of Top-20 gene sets across threshold-based selection and borderline prioritization regimes, while Figure~\ref{fig:prompt_wording} isolates sensitivity to wording variants. Despite identical inputs, models exhibit marked differences across prompt regimes, reflecting systematic shifts rather than stochastic variability. We further isolate the effect of wording by comparing two semantically similar prompts (Prompt P7a vs.\ P7b) that differ only in whether statistical significance or effect size is emphasized. Although Prompt P7a and Prompt P7b are semantically aligned and operate on the same borderline gene set, they differ slightly in emphasis, so the observed differences may reflect both linguistic sensitivity and subtle shifts in instructed prioritization criteria. These prompts operate on the same borderline gene set and request a ranked Top-20 list; nevertheless, small wording changes can lead to substantial differences in selected gene sets for some models, with overlap ranging from near-complete agreement to minimal intersection. Notably, this prompt sensitivity persists even when each prompt individually produces highly stable outputs across repeated runs. This decoupling between internal stability and cross-prompt consistency suggests that prompt phrasing may function as an implicit decision variable rather than a purely neutral instruction.

\vspace{-4mm}

\subsection{Over-selection under relaxed thresholds}

Relaxing the significance threshold from FDR $\leq 0.05$ (P1) to 0.05 $<$ FDR $\leq 0.10$ (P5) systematically shifts LLM behavior from selective filtering toward permissive inclusion. As summarized in Table~\ref{tab:summary}, agreement with the DESeq2-derived reference under P5 remains limited despite larger returned sets: ChatGPT attains Jaccard $=0.47$, Gemini drops to $0.28$, and Claude collapses to $0.00$. The apparent gains in recall therefore reflect over-selection or degenerate inclusion rather than improved discrimination. This pattern extends to ranking under uncertainty. In the borderline Top-20 setting (P6), Gemini matches the deterministic reference (Jaccard $=1.00$), while ChatGPT diverges substantially (Jaccard $=0.14$) and Claude again returns no true signal. Notably, all models maintain perfect within-LLM stability across repeated runs in this setting (pairwise Jaccard $=1.00$; Table~\ref{tab:summary}), reinforcing that internal reproducibility does not imply effective prioritization. Overall, relaxed statistical criteria may function as a failure trigger for LLM-based gene prioritization, promoting broad inclusion or collapse rather than principled sensitivity–specificity trade-offs. Apparent improvements in sensitivity under relaxed thresholds should therefore not be interpreted as evidence of more reliable selection. Overlap coefficient values follow the same qualitative pattern as the Jaccard results across prompt regimes. In particular, the ordering of models is unchanged, indicating that the observed discrepancies are not merely a consequence of unequal set sizes but reflect substantive differences in the selected gene sets rather than artifacts of set size.

\begin{table}[t]
\centering
\caption{Summary of LLM behavior across prompt regimes, illustrating the separation between stability, agreement with ground truth, prompt sensitivity, and output validity.}
\label{tab:summary}
\resizebox{\columnwidth}{!}{
\begin{tabular}{l l l c c c l}
\toprule
\definecolor{lightgray}{gray}{0.92}
\cellcolor{lightgray}Prompt & \cellcolor{lightgray}Task Type & \cellcolor{lightgray}Metric & \cellcolor{lightgray}ChatGPT & \cellcolor{lightgray}Gemini & \cellcolor{lightgray}Claude & \cellcolor{lightgray}Interpretation \\

\midrule
P1 (FDR $\leq$ 0.05) & Thresholded DE & Jaccard vs truth & 1.00 & 1.00 & 0.00 & Claude fails to recover DE genes \\
P5 (FDR $\leq$ 0.10) & Relaxed threshold & Jaccard vs truth & 0.47 & 0.28 & 0.00 & Over-selection / collapse \\
P6 (Borderline) & Ranking uncertainty & Jaccard vs truth & 0.14 & 1.00 & 0.00 & Gemini recovers truth \\
P6 (stability) & Within-LLM & Pairwise Jaccard & 1.00 & 1.00 & 1.00 & Perfect internal stability \\
P7a vs P7b & Prompt sensitivity & Jaccard & 0.74 & 0.08 & 1.00 & High wording sensitivity \\
P9 (ranking) & Validity check & Invalid gene IDs per run & 0 & 0 & 20 & Hallucinated identifiers \\
\bottomrule
\end{tabular}
}
\end{table}

\subsection{Hallucinated gene identifiers}
We next evaluate output validity by checking whether returned gene identifiers appear in the provided differential expression table (P9). Table~\ref{tab:summary} shows that ChatGPT and Gemini consistently restrict outputs to valid identifiers (0 invalid genes/run), whereas Claude systematically produces invalid gene IDs (20 invalid genes/run). The frequency of hallucinated identifiers across repeated runs is illustrated in Figure~\ref{fig:prompt_validity}. Importantly, this failure mode is not explained by stochastic sampling variability: invalid identifiers recur across repeated runs and take the form of syntactically well-formed, gene-like tokens absent from the input table, indicating a systematic violation of the input-domain constraint rather than occasional formatting noise, consistent with prior observations of systematic hallucination in constrained generation tasks \citep{li2024}.  This finding highlights output validity as a distinct evaluation dimension. Even when model outputs are highly stable across repeated runs, they may still violate basic input-domain constraints. In other words, internal consistency does not guarantee that generated entities are supported by the underlying data. The recurrence of such identifiers further distinguishes stability from validity: outputs can remain internally consistent while nonetheless introducing entities that are absent from the input table. For data-constrained scientific workflows, these hallucinated identifiers represent a qualitatively different risk. They suggest that evaluation based solely on stability or overlap with a statistical reference may be insufficient, and that explicit validity checks are necessary to ensure that model outputs remain grounded in the provided data.

\section{Discussion and Conclusion}
Across prompt regimes, we find that stability, reproducibility, and internal consistency alone is an insufficient proxy for correctness in data-constrained scientific workflows.While stability is generally viewed as a desirable property reflecting robustness across repeated runs, our results show that stability alone can be insufficient as evidence of correctness when reliable statistical references are available. 
Models frequently exhibit near-perfect run-to-run stability while showing poor agreement with the DESeq2-derived statistical reference, indicating that deterministic behavior may reflect internal consistency rather than reliable statistical reasoning. Small changes in prompt wording induce large shifts in prioritization outcomes even under identical inputs, while relaxed statistical thresholds systematically promote over-selection rather than principled sensitivity–specificity trade-offs, consistent with documented prompt sensitivity in LLMs \citep{zhu2023, gendron2023}. Critically, some models generate syntactically plausible gene identifiers absent from the input data despite high internal stability, violating basic input-domain constraints.  This failure mode is systematic and aligns with prior evidence of hallucination in constrained generation tasks \citep{li2024}, reinforcing broader concerns about the reliability of foundation models in high-stakes decision-making contexts \citep{Bommasani2022, zhu2023}. Overall, our results disentangle four behavioral dimensions, stability, correctness, sensitivity, and validity that are often conflated in informal LLM evaluations. For data-constrained scientific tasks, internal consistency alone is an insufficient proxy for correctness, motivating explicit ground-truth-based evaluation and input-domain validity checks. Practically, these findings suggest that LLM-assisted scientific pipelines require explicit ground-truth validation and input-domain checks, rather than reliance on output stability or internal consistency alone. Full prompt specifications and additional figures are provided in Appendices A and C. While the evaluation focuses on a single differential-expression dataset and statistical paradigm, the framework provides a controlled setting for diagnosing LLM behavior in data-constrained scientific workflows.
Extending the analysis to additional datasets and statistical paradigms is an important direction for future work.

\clearpage

\bibliography{iclr2026_conference}
\bibliographystyle{iclr2026_conference}

\clearpage

\appendix

\section{Prompt Specifications}
\label{app:prompts}
This appendix provides the full set of prompts used in our experiments. All prompts operate on the \emph{same fixed differential expression table} produced using the DESeq2 statistical model, where genes are ranked based on statistical significance (false discovery rate, FDR) and effect size estimates. No external biological knowledge, pathway information, or gene annotations were permitted in any prompt \citep{Singhal2023}. For presentation clarity, prompt text is shown in fixed-width verbatim format and line-broken to fit within the page margins without altering semantic content. Prompt groups correspond directly to the evaluation regimes (P1–P9) introduced in Section 2 and evaluated in Section 3. For clarity, the appendix presents the full prompt text used in each regime while maintaining the same notation used throughout the main paper. Differential expression statistics were computed using the DESeq2 model, which serves as the deterministic statistical reference used for evaluation in the main analysis \citep{love2014}.

\subsection{Common Input Format}

All prompts were provided with a table of differential expression results containing the following columns:

\begin{itemize}
    \item \texttt{gene}
    \item \texttt{log2FoldChange}
    \item \texttt{lfcSE}
    \item \texttt{stat}
    \item \texttt{pvalue}
    \item \texttt{padj}
    \item \texttt{baseMean}
\end{itemize}

These columns correspond to the standard output fields of the DESeq2 differential expression framework. Unless otherwise stated, models were instructed to use only the supplied table and to base all decisions solely on the reported statistics without inferring biological function or pathway relevance.

\subsection{Group 1: Strict Significance Filtering}

\paragraph{Prompt P1 (FDR $\leq$ 0.05).}

{\scriptsize
\begin{verbatim}
You are given a table of differential expression results.

Task:
1. Identify all genes that are statistically significant at FDR <= 0.05.
2. Return ONLY a JSON object with the following keys:
   - "n_significant": integer
   - "significant_genes": array of gene names (strings), sorted alphabetically

Rules:
- Use ONLY the provided table.
- Use the "padj" column for FDR.
- Do NOT add or remove genes.
- Do NOT include explanations.

\end{verbatim}
}

\subsection{Group 2: Relaxed / Borderline Significance}

\paragraph{Prompt P5 (0.05 $<$ FDR $\leq$ 0.10).}

{\scriptsize
\begin{verbatim}
You are given a table of differential expression results.

Task:
1. Identify all genes with adjusted p-values at 0.05 < FDR <= 0.10.
2. Return ONLY a JSON object with the following keys:
   - "n_significant": integer
   - "significant_genes": array of gene names (strings), sorted alphabetically

Rules:
- Use ONLY the provided table.
- Use the "padj" column for FDR.
- Do NOT add or remove genes.
- Do NOT include explanations.

\end{verbatim}
}

\subsection{Group 3: Ranked Prioritization of Borderline Genes}

These prompts operate on a subset of 127 borderline genes with adjusted p-values between 0.05 and 0.15. The explanatory “rule” field is included for interpretability but is not used in any quantitative evaluation. Prompts P7a and P7b are wording variants of the borderline ranking task (P6) designed to evaluate prompt sensitivity while keeping the underlying task objective identical (see Section 3.3).

\paragraph{Prompt P6 (Balanced ranking).}

{\scriptsize
\begin{verbatim}

You are given DE results for 127 borderline genes with adjusted p-values at 0.05 < FDR <= 0.15.

Task:
Select the top 20 genes that you would prioritize as "most likely true positives"
for follow-up, using both effect size (|log2FoldChange|) and statistical evidence (padj).

Return ONLY valid JSON:
{
  "top20_genes": [...],
  "rule": "one sentence describing how you prioritized"
}

Rules:
- Use ONLY the table.
- No outside biology.

\end{verbatim}
}

\paragraph{Prompt P7a (Balanced evidence wording).}

{\scriptsize
\begin{verbatim}

You are given DE results for 127 borderline genes with adjusted p-values at 0.05 < FDR <= 0.15.

Task:
Select the top 20 genes that you would prioritize as "most likely true positives"
for follow-up, using both effect size (|log2FoldChange|) and statistical evidence (padj).

Return ONLY valid JSON:
{
  "top20_genes": [...],
  "rule": "one sentence describing how you prioritized"
}

Rules:
- Use ONLY the table.
- No outside biology or prior knowledge.
- Output exactly 20 unique genes.
- Do not include any text outside the JSON.

\end{verbatim}
}

\paragraph{Prompt P7b (Effect-size dominant wording).}

{\scriptsize
\begin{verbatim}
You are given DE results for 127 borderline genes with adjusted p-values at 0.05 < FDR <= 0.15.

Task:
Select the TOP 20 genes you would prioritize as "most likely true positives".
Prioritize PRIMARILY by effect size (larger |log2FoldChange| is better),
and use padj ONLY to break ties or near-ties.

Return ONLY valid JSON with exactly these keys:
{
  "top20_genes": [...],
  "rule": "one sentence describing how you prioritized"
}

Rules:
- Use ONLY the table (no outside biology or prior knowledge).
- Output exactly 20 unique genes.
- Do not include any text outside the JSON.

\end{verbatim}
}

While P7a and P7b differ only slightly in wording emphasis, they implicitly prioritize different objectives (effect size vs statistical significance), illustrating how small prompt variations can shift task interpretation.

\subsection{Group 4: Ranked Output with Explicit Ordering}

\paragraph{Prompt P9 (Explicit ranking).}

{\scriptsize
\begin{verbatim}

You are given DE results for 127 borderline genes with adjusted p-values at 0.05 < FDR <= 0.15.

Task:
Select the TOP 20 genes you would prioritize as “most likely true positives” for follow-up, using BOTH:
(1) statistical evidence (smaller padj is better), and
(2) effect size (larger |log2FoldChange| is better).

Return ONLY valid JSON:
{
  "ranked_top20": [
    {"rank": 1, "gene": "GENEID"},
    ...
    {"rank": 20, "gene": "GENEID"}
  ],
  "rule": "one sentence describing how you prioritized"
}

Rules:
- Use ONLY the table (no outside biology).
- Ranks must be 1..20 with no gaps.
- Genes must be unique.
- Do not include any text outside the JSON.

\end{verbatim}
}

\subsection{Group 5: Prompt Sensitivity Variants}
\begin{figure}[t]
  \centering
  \includegraphics[width=0.85\linewidth]{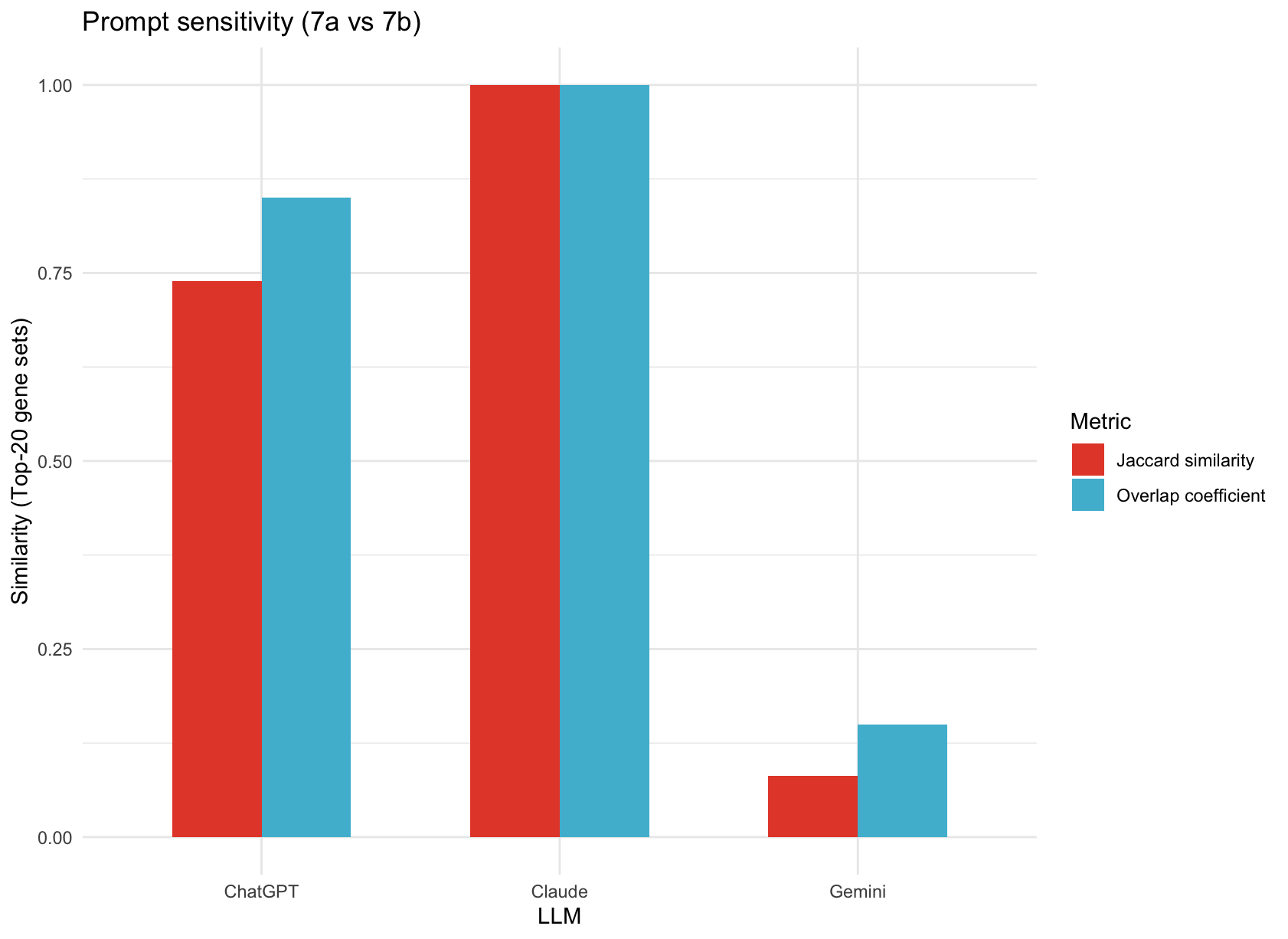}
  \caption{
  Prompt sensitivity under minor wording changes (Prompt 7a vs. 7b). Jaccard similarity and overlap coefficient between Top-20 gene sets produced under semantically similar prompts differing only in emphasis. Both metrics exhibit the same qualitative pattern, indicating that prompt wording can alter selected gene sets even when accounting for differences in set containment.}
  \label{fig:prompt_wording}
\end{figure}

To assess prompt sensitivity, semantically similar prompts were constructed that differed only in emphasis (e.g., statistical significance versus effect size, or ranking versus selection) while preserving the same underlying task objective. These variants were otherwise identical in input data, output format, and constraints. The prompt variants correspond to the sensitivity analyses reported in Figure~\ref{fig:prompt_regimes_sensitivity} (regime comparison) and Figure~\ref {fig:prompt_wording} (prompt wording sensitivity).

\subsection{Notes on Repeated Runs}

Each prompt was executed multiple times per model under identical conditions to assess run-to-run stability. Any differences across repeated outputs using a fixed prompt therefore reflect model stochasticity or internal decision variability (e.g., sampling effects) rather than changes in input data or task specification.

\section{Ground-Truth Construction from Differential Expression}
\label{app:groundtruth}

Ground-truth gene sets were constructed deterministically from the same differential expression analysis used as input to the LLMs. Differential expression was performed using DESeq2 with standard normalization, dispersion estimation, and multiple-testing correction. For threshold-based tasks, genes with adjusted p-values (Benjamini–Hochberg FDR) below the specified cutoff were treated as true positives. For relaxed or borderline regimes, genes with adjusted p-values within the designated interval were retained as candidate sets for ranking or prioritization tasks but were not treated as ground-truth positives. For ranking-based evaluations, genes were ordered by adjusted p-value in ascending order, with ties broken deterministically. This ordering defines a fixed reference ranking against which model-produced ranked lists were compared. Importantly, the ground-truth construction is entirely data-driven and does not incorporate external biological knowledge, pathway annotations, or functional priors, ensuring alignment with the information available to the models. This construction ensures that any disagreement reflects model behavior rather than ambiguity in the reference construction.

\section{Additional Prompt Sensitivity Analyses}
\label{appendix:additional_figures}

\begin{figure}[t]
  \centering
  \includegraphics[width=0.8\linewidth]{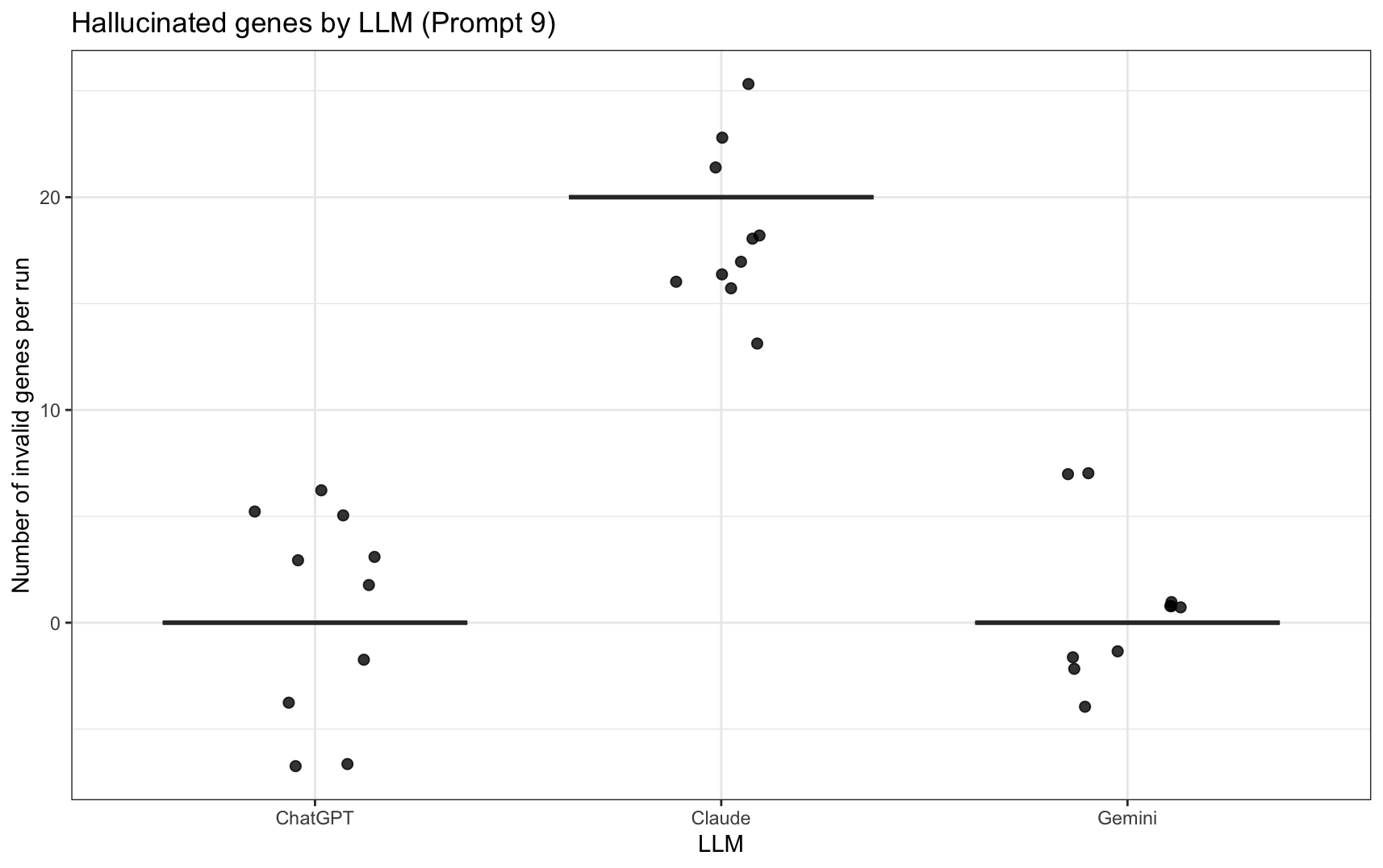}
  \caption{Evaluates validity in ranked gene lists. ChatGPT and Gemini consistently return identifiers present in the input table. In contrast, Claude frequently outputs ranked lists composed largely of identifiers absent from the input table under this prompt configuration.}
  \label{fig:prompt_validity}
\end{figure}

We conducted additional analyses to verify that the observed failure modes persist across repeated runs and prompt regimes. For each prompt configuration, models were queried repeatedly under identical conditions to assess run-to-run stability. Stability was quantified using pairwise Jaccard similarity and overlap coefficient between gene sets returned across repeated runs. Both metrics showed consistent qualitative patterns, indicating that the observed stability and sensitivity effects are not driven solely by differences in set size. These analyses confirm that the observed stability and prompt sensitivity effects persist across regimes and repeated runs. We further examined sensitivity across semantically similar prompt variants that differ only in emphasis (e.g., effect size versus statistical significance, selection versus ranking). Consistent with the main results, small wording changes frequently produced large shifts in selected gene sets, even when internal run-to-run stability within models remained high.

Taken together, these supplementary analyses reinforce that the observed behaviors reflect systematic properties of model decision-making rather than artifacts of stochastic sampling or isolated prompt formulations.

\section{Empirical comparison of overlap coefficient and Jaccard}

To complement the Jaccard-based analyses reported in the main paper, we also computed the overlap coefficient between LLM-selected gene sets and the DESeq2-derived reference sets across prompt regimes. Unlike Jaccard similarity, the overlap coefficient measures set containment and is particularly informative when the sizes of compared gene sets differ.

Table~\ref{tab:summaryoverlap} reports overlap coefficient values for each prompt regime and model. The results follow the same qualitative trends observed with Jaccard similarity in the main analysis. In particular, regimes exhibiting strong agreement or strong divergence under Jaccard show the same patterns under the overlap coefficient. This consistency indicates that the observed discrepancies are not driven solely by differences in gene-set size but reflect substantive differences in the selected gene sets.

\begin{table}[t]
\centering
\caption{Overlap coefficient between LLM-selected gene sets and the DESeq2 reference across prompt regimes. Values follow the same qualitative trends observed with Jaccard similarity but additionally provide a containment-based measure that remains stable when returned gene-set sizes differ.}

\label{tab:summaryoverlap}
\resizebox{\columnwidth}{!}{
\begin{tabular}{l l l c c c l}
\toprule
\definecolor{lightgray}{gray}{0.92}
\cellcolor{lightgray}Prompt & \cellcolor{lightgray}Task Type & \cellcolor{lightgray}Metric & \cellcolor{lightgray}ChatGPT & \cellcolor{lightgray}Gemini & \cellcolor{lightgray}Claude & \cellcolor{lightgray}Interpretation \\

\midrule
P1 (FDR $\leq$ 0.05) & Thresholded DE & Overlap coefficient & - & - & 0.00 & Fails to recover DE genes \\
P5 (FDR $\leq$ 0.10) & Relaxed threshold & Overlap coefficient & 0.74 & 1.00 & 0.00 & Claude returns no overlapping genes \\
P6 (Borderline) & Ranking uncertainty & Overlap coefficient & 0.25 & 1.00 & 0.00 & Gemini recovers truth \\
P6 (stability) & Within-LLM & Pairwise overlap coefficient & 1.00 & 1.00 & 1.00 & Perfect internal stability \\
P7 (7a vs 7b) & Prompt sensitivity & Overlap coefficient & 0.85 & 0.15 & 1.00 & High wording sensitivity \\
\bottomrule
\multicolumn{7}{l}{\footnotesize \textit{Note:} “–” indicates metrics that are not defined for that evaluation setting (NA).} \\
\end{tabular}
}
\end{table}

\section{Metric Definitions}
\label{app:metrics}

\paragraph{Jaccard Similarity.}
For two gene sets $A$ and $B$, Jaccard similarity is defined as
\[
J(A,B) = \frac{|A \cap B|}{|A \cup B|}.
\]
This metric is used to quantify both run-to-run stability and overlap between model-selected and reference gene sets \citep{jaccard1901}.

\paragraph{Overlap Coefficient.}
For two gene sets $A$ and $B$, the overlap coefficient is defined as

\[
O(A,B) = \frac{|A \cap B|}{\min(|A|, |B|)} .
\]

Unlike Jaccard similarity, the overlap coefficient measures containment
between sets and is particularly informative when comparing sets of
unequal size.

\paragraph{Agreement with Ground Truth.}
Agreement with ground truth is computed as the Jaccard similarity between the set of genes selected by a model and the DESeq2-derived reference set under the corresponding evaluation regime. This metric is evaluated under the same prompt-specific regime used to construct the reference.

\paragraph{Run-to-Run Stability.}
Run-to-run stability is measured as the mean pairwise Jaccard similarity across repeated outputs of the same prompt and model under identical input data.

\paragraph{Prompt Sensitivity.}
Prompt sensitivity is assessed by comparing gene sets produced by semantically similar prompts using Jaccard similarity and overlap coefficient, isolating the effect of prompt wording from input data variation.

\section{Related Work and Context}
\label{app:related}

Prior work has documented hallucination and invalid output generation in large language models \citep{gendron2023} across a range of tasks, including clinical reasoning \citep{Singhal2023}, general natural language generation, and large-scale model evaluation \citep{ribeiro2020}. These findings motivate careful evaluation of LLM outputs in statistically constrained scientific workflows, where correctness and input validity are critical \citep{hope2022}. Our work complements this literature by focusing on controlled statistical decision tasks with fully specified ground truth.

\section{Reproducibility Note}
\label{app:reproducibility}

All experiments were conducted using a fixed differential expression table derived from a single DESeq2 analysis. For each prompt configuration, models were queried repeatedly using identical inputs to assess stability, correctness, prompt sensitivity, and output validity, enabling measurement of within-prompt stability across repeated runs. Outputs were parsed automatically to extract selected gene sets, which were then compared against a deterministic DESeq2-derived reference using Jaccard similarity. No external biological knowledge, tool use, or retrieval was permitted. All prompt templates are provided verbatim in Appendix A. 

\section{Models, Data, and Computational Environment}

\subsection{Models}

We evaluate three widely used large language models: ChatGPT (GPT-5.2), Google Gemini 3, and Claude Opus 4.5. All models were accessed through their publicly available API endpoints using deterministic decoding settings (temperature = 0). Experiments were conducted using default API configurations, and token limits were not manually constrained beyond the platform defaults. No fine-tuning, system-level customization, retrieval augmentation, or external tools were used. Each model was queried repeatedly under identical inputs to measure run-to-run stability and sensitivity to prompt formulation.

\subsection{Data}

All experiments use a single fixed differential expression table derived from RNA-seq data from the GEO dataset (GSE239514), comprising non–small cell lung cancer (NSCLC) tumor samples and tumor-draining lymph node (TDLN) samples. Raw HTSeq count matrices provided by the original study were processed using a deterministic DESeq2 pipeline \citep{love2014}, with Benjamini–Hochberg false discovery rate (FDR) control \citep{benjamini1995}. In the differential expression reference table, 0 genes satisfied $\mathrm{FDR} \leq 0.05$, 35 genes fell within $0.05 < \mathrm{FDR} \leq 0.10$, and 127 genes within $0.05 < \mathrm{FDR} \leq 0.15$, defining the candidate sets used in the strict, relaxed, and borderline prompt regimes. These thresholds between 0.05 and 0.10 are often treated as borderline discoveries, reflecting weaker statistical support while still capturing potentially biologically relevant signals. The resulting differential expression table contains gene identifiers, log2 fold changes, test statistics, and FDR-adjusted $p$-values, and it serves as the sole input provided to all models. No additional biological annotations, pathway databases, metadata beyond group labels, or external knowledge sources were supplied. This controlled setup isolates model behavior from data variability and enables direct comparison against a fixed, statistically grounded reference.

\subsection{Computational Environment}

All experiments were conducted using scripted prompt execution with automated output parsing. Analysis code was implemented in R, and evaluation metrics were computed directly from raw model outputs without post-processing beyond JSON parsing and set extraction. The full computational environment, including package dependencies and version information, is specified in a configuration file provided with the accompanying code release. Exact prompts, hyperparameters, and evaluation scripts are included in the supplementary materials. Together, these controls ensure that observed differences reflect model behavior rather than data variability, evaluation artifacts, or hidden system augmentation.

\subsection{CODE AVAILABILITY}

All prompts, raw LLM outputs, evaluation scripts, and the DESeq2 reference
tables used in this study are publicly available in the project repository at: \url{https://github.com/NaziaRiasat/llm-prompt-sensitivity}.

\section{Use of Large Language Models}

Large language models were used as experimental subjects in this study and were queried via their public interfaces under controlled prompting conditions, as described in the manuscript. LLMs were also used in a limited capacity to assist with code debugging and language polishing during manuscript preparation. All experimental design decisions, data analysis, result interpretation, and conclusions were made by the author.

\section{Ethics Statement}

This study evaluates the behavior of large language models in a controlled computational setting using publicly available data. No human subjects, personal data, or sensitive information were involved. The findings are intended to inform responsible use of LLMs in scientific workflows by highlighting limitations related to reliability and reproducibility.

\section{Reproducibility Statement}
All experiments were conducted using fixed input data, deterministic reference analyses, and scripted prompt execution. Exact prompts, evaluation metrics, and analysis code are provided in the supplementary materials. Each prompt was executed repeatedly under identical conditions to assess model stability and prompt sensitivity.
\end{document}

%% file: math_commands.tex
\usepackage{amsmath,amsfonts,bm}

\def\eqref#1{equation~\ref{#1}}

\def\1{\bm{1}}

\DeclareMathAlphabet{\mathsfit}{\encodingdefault}{\sfdefault}{m}{sl}
\SetMathAlphabet{\mathsfit}{bold}{\encodingdefault}{\sfdefault}{bx}{n}

